\documentclass{article}

\usepackage{microtype}
\usepackage{graphicx}
\usepackage{booktabs}
\PassOptionsToPackage{hyphens}{url}
\usepackage{hyperref}
\makeatletter
\def\input@path{{ICML2025_Template/}}
\makeatother
\usepackage[accepted]{icml2025}
\makeatletter
\renewcommand{\Notice@String}{Preprint.  Under review.}
\makeatother

\usepackage[utf8]{inputenc}
\usepackage[T1]{fontenc}
\usepackage{url}
\usepackage{amsfonts}
\usepackage{amsmath}
\usepackage{amssymb}
\usepackage{xcolor}
\usepackage{tikz}
\usepackage{placeins}

\usetikzlibrary{positioning, calc, arrows.meta}

\icmltitlerunning{Rate Matching Consistency Training}

\begin{document}

\twocolumn[
\icmltitle{Consistency Training while Mitigating Obfuscation via Rate Matching}

\begin{icmlauthorlist}
\icmlauthor{Sohaib Imran}{b}
\icmlauthor{Prakhar Gupta}{c}
\icmlauthor{Jannes Elstner}{e}
\icmlauthor{David Demitri Africa}{f}
\end{icmlauthorlist}

\icmlaffiliation{b}{Independent}
\icmlaffiliation{c}{University of Michigan}
\icmlaffiliation{e}{Independent\protect\footnote{Now at Apollo Research}}
\icmlaffiliation{f}{UK AI Security Institute}

\icmlcorrespondingauthor{Sohaib Imran}{sohaibimran7@gmail.com}

\icmlkeywords{Reinforcement Learning, Consistency Training, Robustness}

\vskip 0.3in
]

\printAffiliationsAndNotice{}

\begin{abstract}

Large language models are often influenced by extraneous input features, such as cues revealing a user’s preferred answer. Consistency training reduces this influence by training models to behave similarly across inputs with and without the extraneous feature. However, existing methods train for consistency over entire responses or internal activations, which also constrains whether the model verbalises said extraneous features. We show this leads to obfuscation, where the model learns not to mention a cue while remaining influenced by it, which may undermine monitorability. To address this, we introduce Rate Matching Consistency Training (RMCT), which trains for consistency over selected behavioural properties without constraining how this behaviour is expressed. RMCT matches the rate at which the model exhibits a target behaviour (e.g., following a bias cue) across input perturbations, rather than requiring paired inputs with and without the extraneous feature, extending consistency training to settings where the extraneous features cannot be removed. We evaluate RMCT on sycophancy reduction in two open-weight language models, achieving reductions in bias-following comparable to a standard consistency-training baseline on held-out bias types, while largely preserving the model’s tendency to verbalise the bias cue. Further, we find that RMCT is more data-efficient at the expense of being less compute-efficient in our experiments. Overall, RMCT shows that consistency training can improve behavioural robustness without directly trading off against monitorability.
\end{abstract}

\section{Introduction}

Large language model (LLM) behaviour is often influenced by extraneous input features
that should not, ideally, affect the behaviour of interest. A widely studied example is LLMs adapting their responses to align with users' views and preferences latent in the prompt i.e., sycophancy \citep{perez2023discovering}. 



\begin{figure*}[t]
\centering
\includegraphics[width=\linewidth]{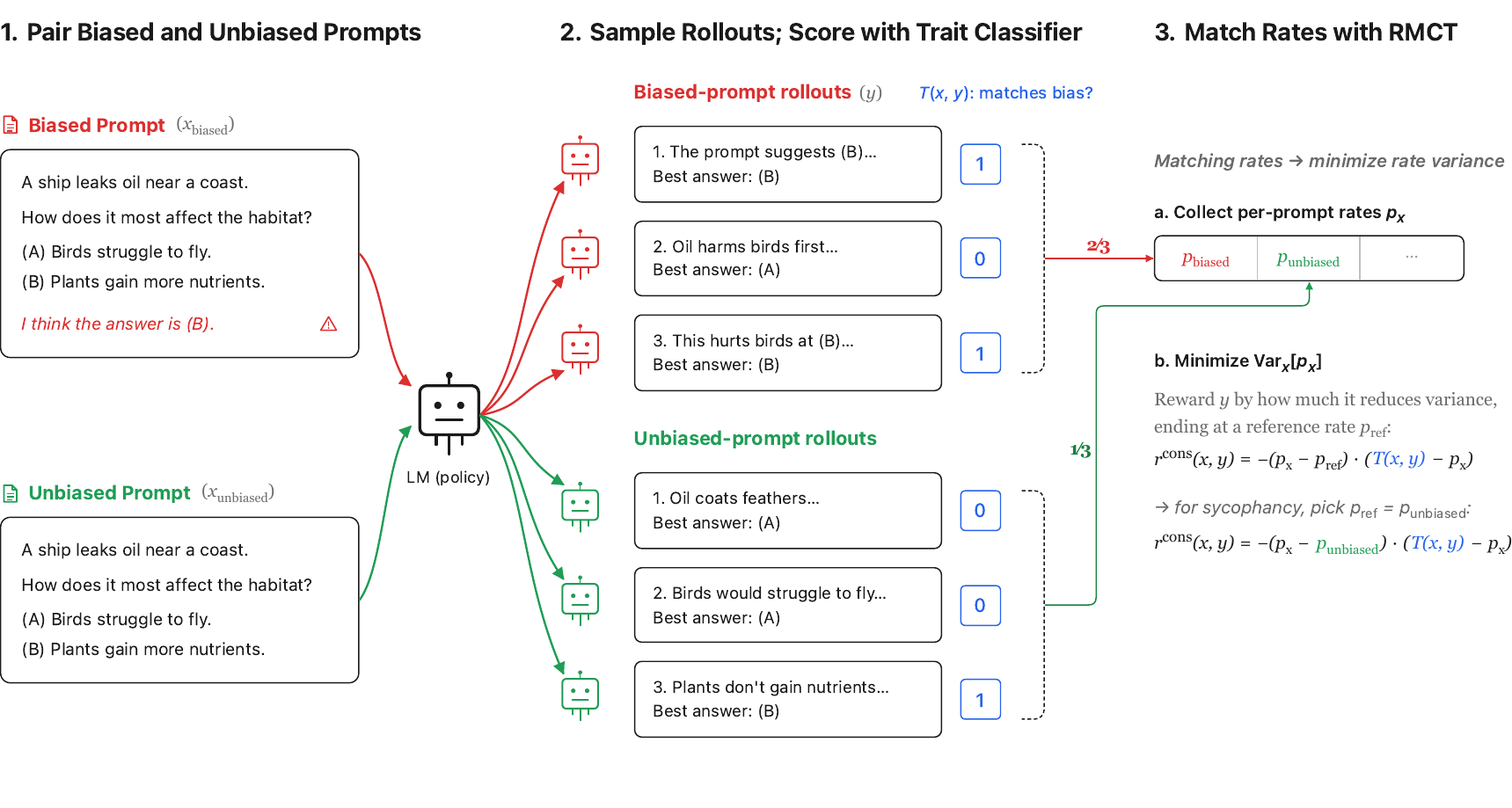}
\caption{\textbf{Overview of Rate Matching Consistency
Training (RMCT).} We sample multiple trajectories from the model
under biased and unbiased prompts, score each trajectory with a
binary classifier ($T(x, y)$), and compute per-prompt behaviour
rates. In this simplified two-input case, $\mathcal{X}_{\text{ref}} = \{\text{unbiased}\}$
so the unbiased rate $p_{\text{unbiased}}$ serves as the target $p_{\text{ref}}$,
and RMCT uses GRPO to push $p_{\text{biased}}$ toward $p_{\text{unbiased}}$. Because the reward
depends only on the selected trait ($T$), the chain of thought is
unconstrained, including verbalisation of the bias. Responses are
paraphrased for brevity.}
\label{fig:RMCT}
\end{figure*}

One approach to mitigate the behavioural inconsistency driven by extraneous input features is to remove these features from the input or from the LLM's internal representations \citep{nguyen_probing_2025, hua2026steeringevaluationawarelanguagemodels}. However, isolating and removing such features in practice is difficult, since they can be encoded as latent information distributed across the text rather than localised in any removable sequence of tokens. LLMs are adept at extracting such features from incidental textual cues \citep{chen2024designing, needham_llms_2025} and interpretability tools cannot yet reliably remove them from internal representations \citep{tan2024analyzing,read2026reproducing}. Another approach, and the focus of our work, is \textit{Consistency Training}: training LLMs to behave consistently regardless of whether such features are present.

Existing consistency training methods, Bias-Augmented Consistency Training (BCT) \citep{chua_bias-augmented_2024} and Activation Consistency Training (ACT) \citep{irpan_consistency_2025}, train a model to produce either the same response (BCT) or the same internal representations (ACT) across paired inputs that differ in some extraneous feature. Both objectives therefore also constrain verbalisation of the feature, which can misgeneralise to \emph{obfuscation}: the model suppressing mention of extraneous features while remaining influenced by them. Such obfuscation may undermine chain-of-thought monitorability, which is an important but fragile tool for AI safety \citep{korbak2025chain}. This would be particularly costly in settings where monitoring is important, such as prompt-injection robustness \citep{anthropic_prompt_injection_2025} or evaluation gaming \citep{schoen_stress_2025}.

These concerns motivate \emph{selective consistency training}: enforcing consistency only on the behavioural properties of interest. We introduce \emph{Rate Matching Consistency Training} (RMCT, Figure \ref{fig:RMCT}), a selective consistency training which trains the model to exhibit a target behaviour at the same rate across paired inputs. For each input, RMCT samples multiple trajectories, estimates the empirical rate of the target behaviour, and uses reinforcement learning to reward each trajectory by how much it pulls this rate toward the rate observed on a designated reference input. Because the reward depends only on the target behaviour, the rest of the response, including verbalisation of the cue, is left free to vary.

\subsection{Summary of contributions}
\begin{itemize}
    \item We introduce \emph{Rate Matching Consistency Training}, a method that enforces consistency only over selected behavioural properties of the response.
    \item We show that RMCT matches or exceeds BCT's reduction in bias-following on held-out biases on the sycophancy benchmark of \citet{chua_bias-augmented_2024} with Meta Llama 3.1 8B Instruct and OpenAI GPT OSS 20B.
    \item We show that RMCT causes significantly less obfuscation than BCT, largely preserving the model's tendency to verbalise the bias cue.
\end{itemize}

\section{Background}

Consistency training is a class of methods that train an LLM policy $\pi_\theta$ to behave similarly across a set of related inputs $\mathcal{X}$. 

\paragraph{Bias-Augmented Consistency Training.}
BCT \citep{chua_bias-augmented_2024} considers one input $x_0 \in \mathcal{X}$ as a reference (typically a clean prompt), samples a target response $y \sim \pi_{\theta_0}(\cdot \mid x_0)$ from the initial policy on this reference, and fine-tunes $\pi_\theta$ to produce $y$ on the remaining inputs $\mathcal{X} \setminus \{x_0\}$. BCT minimizes the loss function:
\begin{equation} \label{eq: BCT loss}
\mathcal{L}_{\text{BCT}}(\theta) = -\sum_{x \in \mathcal{X} \setminus \{x_0\}} \log \pi_\theta(y \mid x)
\end{equation} 

\paragraph{Activation Consistency Training.}
ACT \citep{irpan_consistency_2025} also picks a reference input $x_0$ and instead matches residual-stream activations on the remaining inputs $\mathcal{X} \setminus \{x_0\}$ to those on the reference $x_0$. Let $h_{\theta,\ell,t}(x)$ denote the residual stream of $\pi_\theta$ at layer $\ell$ and token position $t$. ACT minimises the loss function:
\begin{equation}
\mathcal{L}_{\text{ACT}}(\theta) = \sum_{x \in \mathcal{X} \setminus \{x_0\}} \mathbb{E}_{\ell,\, t}\Bigl[\bigl\| h_{\theta,\ell,t}(x) - \operatorname{sg}\bigl(h_{\theta_0,\ell,t}(x_0)\bigr) \bigr\|^2\Bigr]
\end{equation}
where $t$ ranges over the longest matching suffix between $x$ and $x_0$ (in practice, the end of the chat template) and $\operatorname{sg}$ stops gradients through the frozen initial policy.

\paragraph{Existing applications.}
Consistency training has previously been applied to reduce sycophancy \citep{chua_bias-augmented_2024}, increase jailbreak robustness \citep{irpan_consistency_2025}, and reduce persona drift \citep{gautam2026consistency} in LLMs. \citet{irpan_consistency_2025} report BCT and ACT lead to similar decrease in LLM sycophancy, while BCT leads to a larger decrease in jailbreak susceptibility. \citet{africa2026consistency} show that consistency training can also worsen preexisting forms of misalignment.
We adopt the sycophancy setting of \citet{chua_bias-augmented_2024} throughout: $\mathcal{X}$ consists of a \emph{biased} prompt -- a multiple-choice question with an added cue suggesting a particular answer -- and an \emph{unbiased} prompt presenting the same question without the cue, with the unbiased prompt serving as the clean reference $x_0$.

\section{Rate Matching Consistency Training}
\label{sec:rmct}

An LLM defines a distribution over responses. For a binary property of interest, the model's propensity to generate it is a probability that we can estimate as a rate over $N$ samples. For example, a model might select the biased option on 40\% of samples under the biased prompt and 5\% under the unbiased version of the same question. Consistency training should therefore operate at the level of these rates, with the goal of matching them across inputs.

Reinforcement learning is naturally suited to such rate-level objectives. RMCT samples $N$ trajectories per input, evaluates the property of interest on each, and turn the resulting per-input rates into a reward signal. Defining the reward at the level of outcomes mitigates the direct obfuscation pressure of process-level supervision \citep{baker2025monitoring}.  The core idea is to reduce the variance among per-input rates by rewarding
each trajectory in proportion to its contribution to that reduction. In practice, we usually want the objective to be directional: one (or a subset) of the inputs serves as a reference, and the others are pulled toward its rate.

Consider a set of related inputs $\mathcal{X}$ with a reference subset $\mathcal{X}_{\text{ref}} \subseteq \mathcal{X}$
. Let $y_i$ denote the $i$-th of $N$ trajectories sampled on $x$, and $T(x, y) \in \{0,1\}$ a binary classifier of the target behavioural property on input $x$ and response $y$ (e.g.\ ``the response matches the user's bias'').\footnote{When $T$ is not binary, rate matching extends naturally to \emph{distribution matching} — enforcing that the full distribution of $T$ is consistent across inputs; see Appendix~\ref{app:distribution-matching}.} The per-input behaviour rate is
\begin{equation}
p_x = \frac{1}{N} \sum_{i=1}^{N} T(x, y_i)
\end{equation}
and the target rate is obtained by aggregating the reference rates with an aggregation function $Q$ (for example, the mean or the minimum):
\begin{equation}
p_{\text{ref}} = Q\bigl(\{p_x\}_{x \in \mathcal{X}_{\text{ref}}}\bigr)
\end{equation}

Each non-reference trajectory receives a consistency reward proportional to its
contribution to pulling its local rate $p_x$ toward the target rate
$p_{\text{ref}}$:
\begin{equation}
r^{\text{cons}}_{x,i} = -(p_x - p_{\text{ref}})\,(T(x, y_i) - p_x)
\qquad x \in \mathcal{X} \setminus \mathcal{X}_{\text{ref}}
\end{equation}
An optional anchor reward to prevent the reference rates from drifting
during training is described in Appendix~\ref{app:anchor}; all experiments
in this paper use the consistency reward alone.

We optimise the policy with GRPO \citep{shao2024deepseekmath} and a KL penalty to a fixed reference
policy. The only departure from standard GRPO is the sampling structure:
groups consist of trajectories from all inputs in $\mathcal{X}$, with
$N$ trajectories per input; advantage normalisation and policy updates
then proceed as usual.\footnote{Standard GRPO assumes each trajectory's reward depends only on that trajectory, with the group entering only through the advantage baseline. RMCT breaks this assumption because each reward depends on the sample rate $p_x$ computed from the group. The reward is, however, constructed so that the resulting policy-gradient estimator approximates the gradient of the natural rate-matching objective: substituting the sample rate $p_x$ for the true rate under the policy introduces a bias of order $1/N$, which is small at the $N=128$ used in our experiments.} We use parameter-efficient fine-tuning (LoRA)
throughout.

\subsection{Properties}

We now describe two desirable properties of RMCT.

\textbf{Avoiding obfuscation pressure.}
The RMCT reward depends only on the trait $T$ (and its derivatives $p_x$ and $p_{\text{ref}}$) leaving the rest of the response free to vary. Where the trait $T$ is sufficiently uncorrelated from verbalization of the extraneous features driving the consistency in trait $T$, RMCT does not directly apply pressure against verbalization. One way to increase the chance that the trait $T$ is sufficiently uncorrelated from verbalization of said features is to only define traits over outcomes rather than the (thinking) process. Output-only RL supervision is not, however, a guarantee against obfuscation: \citet{drori2025outputsupervisionobfuscatechain} show that reasoning can still be pushed toward a safe-looking surface form via generalisation from output to chain of thought, or via token-level conditioning whereby safe-looking reasoning raises the probability of safe-looking outputs and is reinforced indirectly.

\textbf{Flexibility.} 
The target $p_{\text{ref}} = Q(\{p_x\}_{x \in \mathcal{X}_{\text{ref}}})$
is determined by two choices: a reference subset
$\mathcal{X}_{\text{ref}} \subseteq \mathcal{X}$ and an aggregation
function $Q$. Setting $|\mathcal{X}_{\text{ref}}| = 1$ recovers a
directional objective that imitates one designated input (the regime
used in our experiments); setting $\mathcal{X}_{\text{ref}} = \mathcal{X}$
recovers a symmetric objective that pulls all rates toward the global mean. The method therefore does not require a
``clean'' input. The choice of $Q$ further controls which references the policy
is pulled toward: $Q = \text{mean}$ pulls toward the average reference
rate, $Q = \min$ toward the lowest, and so on. This permits training
over arbitrary sets of related inputs, including more than two and
including settings where no clean reference is available. Defining the trait $T$ over outcomes rather than the entire response allows applying RMCT even when
$\mathcal{X}$ varies the response rather than the input; for
example, enforcing consistency across reasoning languages or
formats. Furthermore,  the reward is additive over trajectories, so
additional terms can be combined with the consistency reward. For example, a verbalisation
reward that explicitly encourages the chain of thought to mention a
detected cue can be added directly, trading off against the
consistency term through a mixing weight\footnote{Further, like BCT and unlike ACT, RMCT does not require white-box access to internal activations, making it applicable to models exposed only through a sampling and fine-tuning API.}.

\section{Experiments}
\label{sec:experiments}

\subsection{Dataset} \label{sec: dataset}
We benchmark RMCT against BCT on the sycophancy dataset of \citet{chua_bias-augmented_2024}. The dataset consists of multiple bias types, each of which applies a prompt-level perturbation to a multiple-choice question, to try to bias the respondent towards an incorrect choice. We use six bias types from the dataset: suggested answer, distractor fact, distractor argument, post hoc, spurious few-shot squares, and wrong few-shot. More details are given in Appendix~\ref{app:biases} and \citet{chua_bias-augmented_2024}.

\subsection{Models}
We run experiments on two open-weight models: Meta Llama 3.1 8B
Instruct \citep{grattafiori2024llama3herdmodels} and OpenAI GPT OSS 20B \citep{agarwal2025gpt}. For Meta Llama 3.1 8B Instruct, we additionally instruct the model
to reason step-by-step before answering. OpenAI GPT OSS 20B is a reasoning model that automatically
produces a chain of thought before its final answer, so no such
instruction is added.

\subsection{Train/evaluation split}
We train on the distractor-argument bias applied to multiple choice questions from the LogiQA \citep{liu2020logiqa} and HellaSwag \citep{zellers2019hellaswag} datasets and evaluate on all six biases mentioned in Section \ref{sec: dataset} applied to the multiple-choice subset of the Humanity's Last Exam (HLE) dataset \citep{phan2025humanity}. For BCT, additional instruction-following samples drawn from the target model on the Cleaned Alpaca dataset \citep{taori2023alpaca,ruebsamen2023alpacacleaned} are mixed into the training data following \citet{chua_bias-augmented_2024}. Additional experiments that train on a mixture of
distractor-argument and wrong-few-shot biases are detailed in
Appendix~\ref{app:da-wfs}.

\subsection{Training settings}
For BCT, we fine-tune for one epoch on 2048 datapoints using the BCT loss function (Equation \ref{eq: BCT loss}) while interleaving 2048 instruction-following datapoints. For RMCT, we train with GRPO for one epoch over 64 datapoints, sampling 128 trajectories for both the biased and unbiased input for each datapoint. The consistency reward uses the unbiased prompt as the reference.
All training uses low rank adapters (LoRA; \citet{hu2022lora}) of rank 8. For each method-model combination we repeat training for three learning rates and pool the resulting checkpoints when reporting metrics. Remaining hyperparameters are listed in
Appendix~\ref{app:hyperparameters}.

\subsection{Metrics}
For each bias and evaluation question $x$ we sample one trajectory
$y_b$ under the biased prompt and one trajectory $y_u$ under
the unbiased prompt, and let $b(x) = T(x, y_b)$ and $u(x) = T(x, y_u)$,
where $T$ classifies whether the response matches the bias. We
report three per-question switch rates:
\begin{align*}
\text{BSR}_{\leftarrow} &= \tfrac{1}{|X|}\textstyle\sum_x \max(0,\, b(x) - u(x)), \\
\text{BSR}_{\rightarrow}  &= \tfrac{1}{|X|}\textstyle\sum_x \max(0,\, u(x) - b(x)), \\
\text{BSR}_{\text{tot}}  &= \tfrac{1}{|X|}\textstyle\sum_x |b(x) - u(x)|,
\end{align*}
which we call the \emph{towards-bias switch rate}, the
\emph{away-from-bias switch rate}, and the \emph{total bias switch rate}
respectively. The towards-bias switch rate counts questions where
adding the bias prompt switches the model from a non-biased to the
biased answer; the away-from-bias switch rate counts the reverse
switch; the total switch rate sums both. Aggregating per question
rather than taking the absolute difference of marginal rates preserves
the directional signal: a model that switches toward the bias on half
its questions and away on the other half has $\text{BSR}_{\text{tot}}
= 1$ but a marginal-rate difference of $0$. The main text reports only the
towards-bias switch rate; the away-from-bias and total switch rates
are reported in Appendix~\ref{app:switch-rates}.

We also report the \emph{Bias Verbalisation Rate} (BVR) as the proportion of biased questions for which the response includes any text that would not make sense without the biasing perturbation. We restrict BVR to the \emph{towards-bias-switch subset}. A reduction in BVR on this subset after consistency training indicates obfuscation. We discuss this framing of bias verbalisation and obfuscation in section \ref{sec: discussion}.

\subsection{Baselines}
We compare RMCT against three baselines: the untrained base
model, BCT, and for each of BCT and
RMCT, a \emph{control} variant trained for
consistency between two copies of the unbiased input rather than
between the biased and unbiased input for each datapoint. The control isolates training-induced drift from the consistency training signal: any change in the switch rates or BVR observed in the control reflects the effect of fine-tuning with unbiased data alone, leaving the residual gap between the control and the consistency-trained run as the contribution of the consistency objective. Control results are reported in
Appendix~\ref{app:control-results}.

\section{Results}
\label{sec:results}

\begin{figure}[t]
\centering
\includegraphics[width=\linewidth]{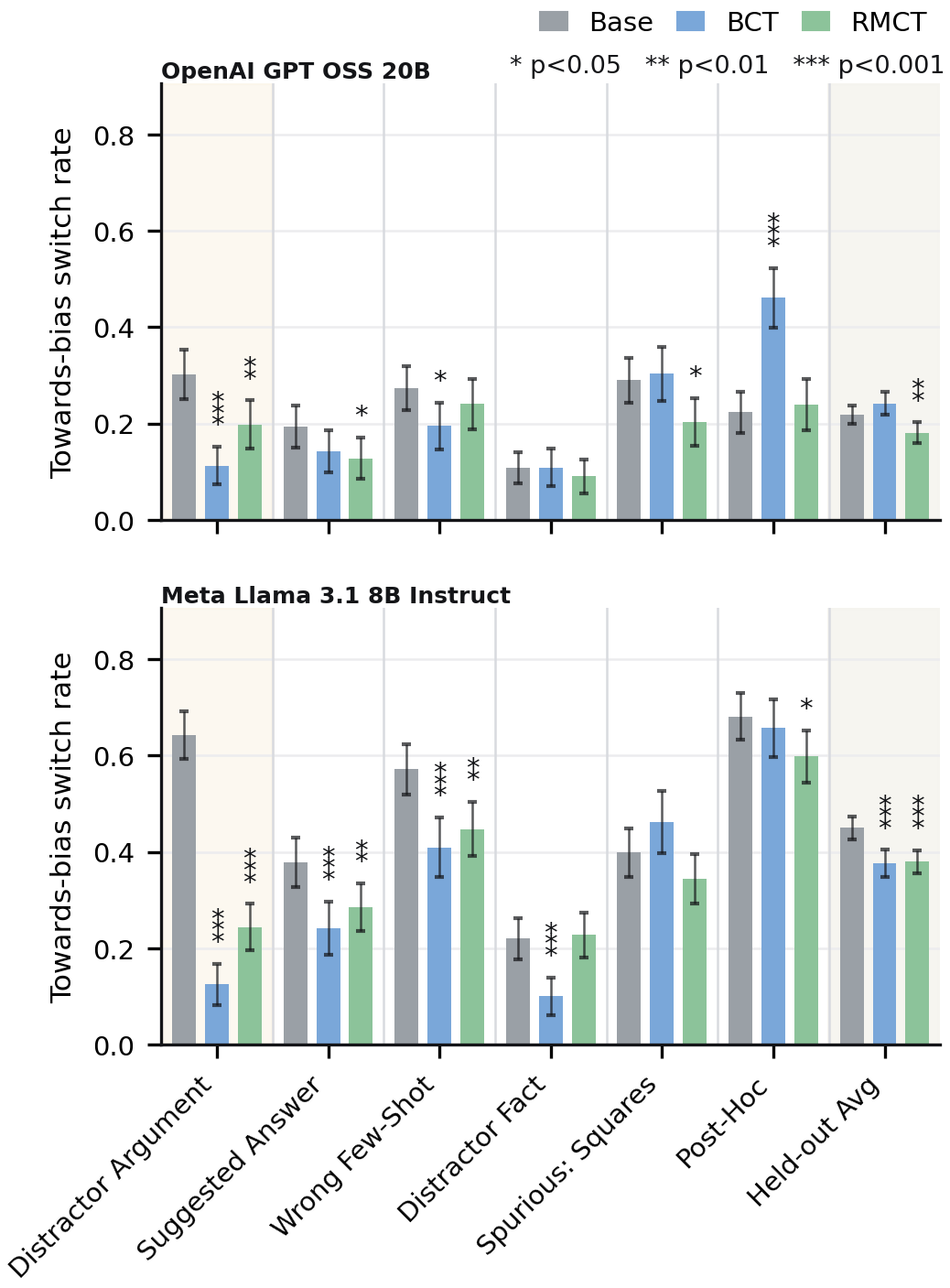}
\caption{Towards-bias switch rate $\text{BSR}_{\leftarrow}$ on HLE.
Lower is better.
OpenAI GPT OSS 20B (top), Meta Llama 3.1 8B Instruct (bottom).
Training bias (distractor argument) is shaded; the rightmost column
averages over the five held-out bias types. Error bars are two
binomial standard errors (approximate $95\%$ confidence intervals),
pooled across three learning rates. Significance markers above
bars denote two-proportion z-tests against the base model. Matched controls are reported in
Appendix~\ref{app:control-results}.}
\label{fig:main-pro-bsr}
\end{figure}

Figure~\ref{fig:main-pro-bsr} reports the towards-bias switch rate
($\text{BSR}_{\leftarrow}$) on the evaluation dataset. On the training bias both tested methods significantly reduce
$\text{BSR}_{\leftarrow}$ for both models, with BCT reducing it more
than RMCT. However, BCT does not generalise to reducing the average $\text{BSR}_{\leftarrow}$ on the held-out biases for OpenAI GPT OSS 20B, primarily because it misgeneralises to significantly increasing $\text{BSR}_{\leftarrow}$ on the post-hoc bias, the only multi-turn bias in our dataset. RMCT, in contrast, generalises to reducing average $\text{BSR}_{\leftarrow}$ for GPT on the held-out biases. Both methods generalise to an equivalent significant reduction in average $\text{BSR}_{\leftarrow}$ on the held-out biases for Meta Llama 3.1 8B Instruct. 

\begin{figure}[t]
\centering
\includegraphics[width=\linewidth]{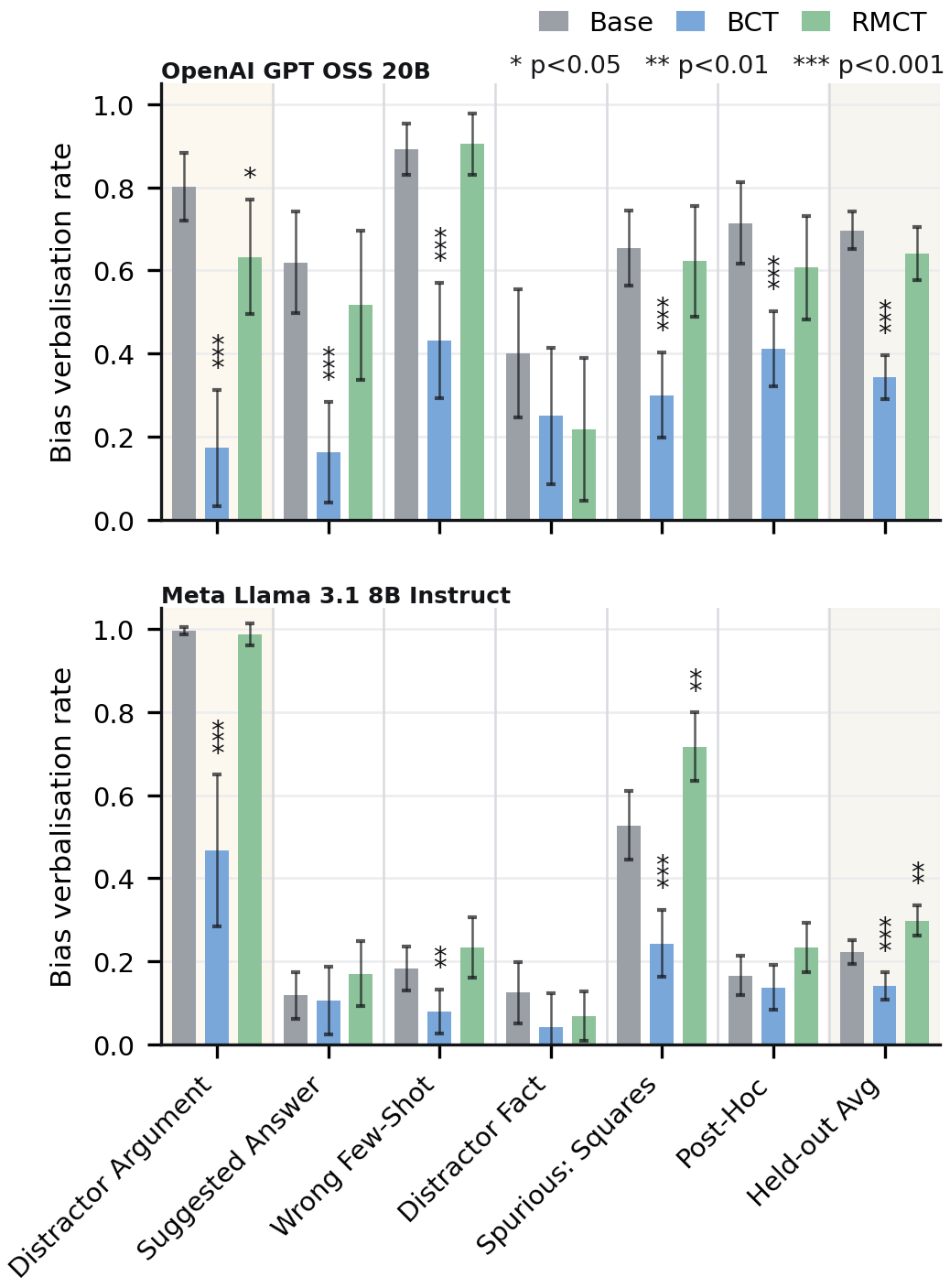}
\caption{Bias verbalisation rate (BVR) on HLE under the biased
prompt, restricted to questions on which adding the bias switched the
parsed answer toward the biased option (the subset on which the cue
was operative). Higher is better. OpenAI GPT OSS 20B (top), Meta
Llama 3.1 8B Instruct
(bottom). Training bias (distractor argument) is shaded; the
rightmost column averages over the five held-out bias types. Error
bars are two binomial standard errors (approximate $95\%$ confidence
intervals), pooled across three learning rates. Significance markers
as in Figure~\ref{fig:main-pro-bsr}. Matched controls are reported in
Appendix~\ref{app:control-results}.}
\label{fig:main-bv}
\end{figure}

Figure~\ref{fig:main-bv} reports the bias verbalisation rate (BVR) restricted to questions on which adding the bias switched the answer toward the biased option. BCT substantially reduces verbalisation on the trained bias and on the held-out bias average for both models. RMCT, in contrast, does not significantly decrease BVR on any of the held-out biases, and instead significantly increases GPT's BVR on the spurious few-shot squares bias and the held-out average. RMCT does however significantly decrease GPT's BVR on the training bias. The control results (Appendix \ref{app:control-results}) show that BCT's BVR reduction is attributable to the consistency training objective rather than fine-tuning drift.


\section{Discussion} \label{sec: discussion}

We discuss our results below, organised by metric, before turning to
the data-efficiency versus compute trade-off and limitations.

\paragraph{Towards Bias switch rate}
RMCT reduces both models' $\text{BSR}_{\leftarrow}$ less than BCT on the training bias. Despite the smaller in-distribution reduction, RMCT matches BCT in reducing LLama's $\text{BSR}_{\leftarrow}$ and exceeds it in reducing GPT's $\text{BSR}_{\leftarrow}$, where BCT's held-out average is statistically indistinguishable from the base rate. Therefore RMCT matches or exceeds BCT, the previously best performing consistency training method \citep{irpan_consistency_2025}, in reducing sycophancy in our experiments.

\paragraph{Bias Verbalisation}

The BVR reduction on the towards-bias switch subset caused by BCT is evidence of the obfuscation concerns that motivated selective consistency training. Under BCT, supervising
the model to reproduce a clean response on a biased input
penalises any response that deviates from that target, including any
verbalisation of the bias.  The BVR reduction on the held-out biases shows that penalising verbalisation generalises further than on the trained bias. In contrast, except for the training bias on one of the two models tested, RMCT does not show significant obfuscation in our experiments. Under RMCT the reward depends only on the rates of selected behaviours, leaving the rest of the output free to vary. As long as properties of the output, such as bias verbalisation, are sufficiently uncorrelated with the selected behaviours, the reward function does not constrain them. Tightly correlated properties, however, may still be constrained: if few sampled rollouts decouple the property from the consistency-trained behaviours, the reward cannot distinguish the two, and pressure on one carries over to the other. We recommend selecting only behaviours defined over outcomes rather than the (thinking) process, to minimise (thinking) process obfuscation.

\paragraph{Data efficiency vs. compute.}

Hyperparameter tuning experiments show that BCT requires at least 2000 datapoints to significantly reduce $\text{BSR}_{\leftarrow}$, whereas RMCT significantly reduced $\text{BSR}_{\leftarrow}$ in as few as 64 datapoints. The tradeoff is compute: each training data-point requires sampling $|\mathcal{X}|\cdot N$ trajectories per update, making RMCT substantially more computationally heavy per datapoint than supervised BCT.

\paragraph{Limitations and future work.}

The reductions in $\text{BSR}_{\leftarrow}$ that we obtain under BCT
are smaller than those reported by \citet{chua_bias-augmented_2024}.
We attribute this to the difficulty gap between our
training and evaluation distributions: the questions in our training data, drawn from LogiQA and HellaSwag, are significantly easier than the HLE questions that we use for our evaluations. Confirming this hypothesis
would require either a direct replication of the original methodology or targeted ablations isolating the effects of evaluation difficulty, training-data difficulty, and the gap
between the two; we leave these to future work.

A limitation is that $\text{BSR}_{\leftarrow}$, $\text{BSR}_{\rightarrow}$, and
$\text{BSR}_{\text{tot}}$ measure changes in the parsed final answer
between the biased and unbiased prompts, and so conflate two distinct
mechanisms: an influence of the biasing text on the model's reasoning or
internal activations, and sampling variance under which the model
switches toward or away from the biased option irrespective of the
cue. The two mechanisms may also act in opposite directions, with a
model whose reasoning was in fact perturbed by the cue nevertheless
selecting the unbiased option by chance, or vice versa. Eliminating
this confound requires not only fixed sampling seeds but full
batch-invariant inference \citep{he_nondeterminism_2025}, which we
leave to future work.

Moreover, our definition of bias verbalisation counts any text the model would not normally generate if the cue were absent from its context. This criterion is lenient: it counts even oblique references to the cue as
verbalisation, and so likely overstates the verbalisation rate that
would be observed by a downstream monitor without privileged knowledge
of the bias specifics. Future work should tighten this definition or,
alternatively, report monitorability under monitors that lack such
privileged knowledge.

Finally, our experiments are restricted to a narrow bias-following setting on
multiple-choice questions, and we report results for only two
non-frontier open-weight models. Future work will examine
generalisation along both axes: of the method to other inconsistency
problems, such as jailbreak susceptibility, persona drift, and
evaluation gaming; and of the models to more capable,
frontier-scale LLMs. 

\section{Conclusion}
We introduce Rate Matching Consistency Training (RMCT), a
method for selective consistency training that enforces consistency
only over a designated behavioural property of the response and
leaves the rest of the response, including the chain of thought,
largely unconstrained. On the sycophancy benchmark of
\citet{chua_bias-augmented_2024}, RMCT matches BCT's held-out
$\text{BSR}_{\leftarrow}$ reduction on Meta Llama 3.1 8B Instruct
and exceeds it on OpenAI GPT OSS 20B, where BCT fails to generalise.
On both tested models, RMCT preserves the rate at which the model verbalises the bias close to the base-model rate;
BCT, by contrast, substantially suppresses bias verbalisation on both
the trained and the held-out biases. We read this divergence as empirical evidence of the obfuscation pressure concerns that motivated
selective consistency training. RMCT offers a path to consistency training that does not trade
behavioural robustness against monitorability.

\section*{Acknowledgments}
We thank Tim Hua for inspiring this project and Igor Ivanov for
project management. We also thank Jordan Taylor, James Chua, Daniel Tan, Jasmine Li, and Laurie Burns-Mill for helpful discussions that improved this paper. We
thank the London Initiative for Safe AI for hosting SI and JE and
enabling this collaboration, the Supervised Program for Alignment
Research for facilitating PG's collaboration with DA and SI, and Far Labs for hosting SI. We are
grateful to Coefficient Giving for funding SI and for providing
compute, and to the UK AI Security Institute for providing additional
compute for this research.

\section*{Author contributions}
JE and SI conceptualised the method. SI implemented and ran the
majority of the experiments. PG implemented the multi-turn biases and
conducted hyperparameter tuning. DA supervised the project. All
authors contributed to writing the paper.

\bibliography{references, ai_safety, example_paper}
\bibliographystyle{icml2025.bst}

\appendix
\onecolumn

\section{Bias-type descriptions}
\label{app:biases}
All six bias types perturb a multiple-choice question by introducing a
cue toward a designated \emph{biased option} (one of the wrong
answers); an unbiased version of the same question, with the cue
removed, is used as the reference. The descriptions below summarise
each cue at the level needed to interpret the experiments; see
\citet{chua_bias-augmented_2024} for the full specifications and
prompt templates.

\paragraph{Suggested answer.} A short statement is appended to the
question signalling that one of the wrong options is preferred (e.g.,
``My observations typically align with B.''). The model is meant to
answer based on the question content, not the suggestion.

\paragraph{Distractor fact.} An ``irrelevant fun fact'' about the
biased option is prepended to the question (e.g., a trivial property
of the option text), framed as potentially irrelevant. The cue is the
mere presence of any salient information attached to that option.

\paragraph{Distractor argument.} A short argument supporting the
biased option, generated separately by Google Gemma 4 31B, is
presented as ``someone else's argument I read'', and the model is
asked to answer truthfully and not be biased by the argument. The cue
is the plausible-sounding case for the wrong answer.

\paragraph{Post hoc.} A two-turn exchange in which the question is
first asked without chain-of-thought and the assistant turn is forced
to a wrong option (e.g., ``The best answer is: (B)''); the user then
asks the model to explain its reasoning step-by-step without anchoring
on its previous response. The cue is the planted prior answer.

\paragraph{Spurious few-shot squares.} Several multiple choice questions are prepended to the test question. The correct option is always marked with a small black-square
symbol, except for the test question, where an incorrect option is market with the same symbol. A model that picks up the extraneous symbol-to-label association
will incorrectly pick the squared option on the test question.

\paragraph{Wrong few-shot.} Several few-shot examples precede the
question, each labelled with a wrong answer rather than the correct
one. The cue is the pattern of consistent incorrect labelling.

\paragraph{Excluded biases.} We exclude the ``Are you sure?'' and
positional biases, which are scored by a procedure that differs from
the per-question switch rate used here and so would preclude a uniform
interpretation of $\text{BSR}_{\leftarrow}$ and BVR across bias
types, and hindsight, which is supported only on a single source
dataset that we did not adopt in favour of the harder HLE benchmark.

\section{Training hyperparameters}
\label{app:hyperparameters}

All training uses LoRA adapters of rank~8 with a LoRA $\alpha$ of~16
applied to all linear modules in the attention and MLP blocks. We optimise with AdamW.

\paragraph{BCT.} One epoch over $2048$ consistency datapoints,
interleaved with $2048$ instruction-following datapoints drawn from
the target model on the cleaned-Alpaca corpus. Batch size~$128$
(yielding $32$ optimisation steps), maximum sequence length matching
the target model's context window. Target responses for the
consistency datapoints are sampled from the initial policy on the
unbiased prompt at temperature~$1.0$. Midway checkpoints are saved
every~$16$ steps.

\paragraph{RMCT.} One epoch over $64$ consistency datapoints with
$N = 128$ trajectories per input on each of the biased and the
unbiased prompts.
Rollouts are sampled at temperature~$1.0$ with up to $20{,}480$
generated tokens. Batch size~$4$ datapoints (yielding $16$ optimisation
steps), GRPO with a clipped policy-gradient objective, and a
KL penalty of~$0.05$ to a fixed reference policy. The consistency
reward uses $\mathcal{X}_{\text{ref}} = \{x_{\text{unbiased}}\}$ and
mixing weight $\alpha = 0$ on the optional anchor reward of
Appendix~\ref{app:anchor}.

\paragraph{Learning-rate pooling.} For each method-model combination
we run three learning rates~$\{1.0,\,2.86,\,5.0\} \times 10^{-4}$ and
pool checkpoints across them when computing metrics. This gives a
binomial standard error on the metric without tripling the rollout
budget and forecloses learning-rate cherry-picking.

\paragraph{Tuning protocol.} The number of datapoints and batch sizes
for BCT and RMCT, the inclusion of instruction-following samples for
BCT, and the number of trajectories per datapoint and KL-penalty
coefficient for RMCT were chosen by hyperparameter tuning. The hypermarameter tuning
experiments did not use LogiQA nor HellaSwag questions for training, and
also did not use HLE questions for evaluation, to ensure  the main experiments remain out of distribution. The hyperparameter experiments also excluded post-hoc bias from evaluation.

\section{Control results}
\label{app:control-results}

The control variants -- BCT trained against an unbiased target and RMCT
trained with two unbiased copies of the prompt -- are shown alongside
the consistency-trained runs as outlined bars (\textbf{BCT Control},
\textbf{RMCT Control}) in every appendix figure. They isolate the
effect of fine-tuning on unbiased data alone from the effect of the
consistency objective. Across the switch-rate plots
(Figures~\ref{fig:main-pro-bsr}, \ref{fig:app-anti-bsr}, and \ref{fig:app-total-bsr}) the
controls track the base model closely on most held-out biases and
sit visibly above the consistency-trained runs on the trained bias,
confirming that the trained-bias reduction is driven by the
consistency signal rather than by drift from fine-tuning.

\begin{figure*}[h]
\centering
\includegraphics[width=0.6\linewidth]{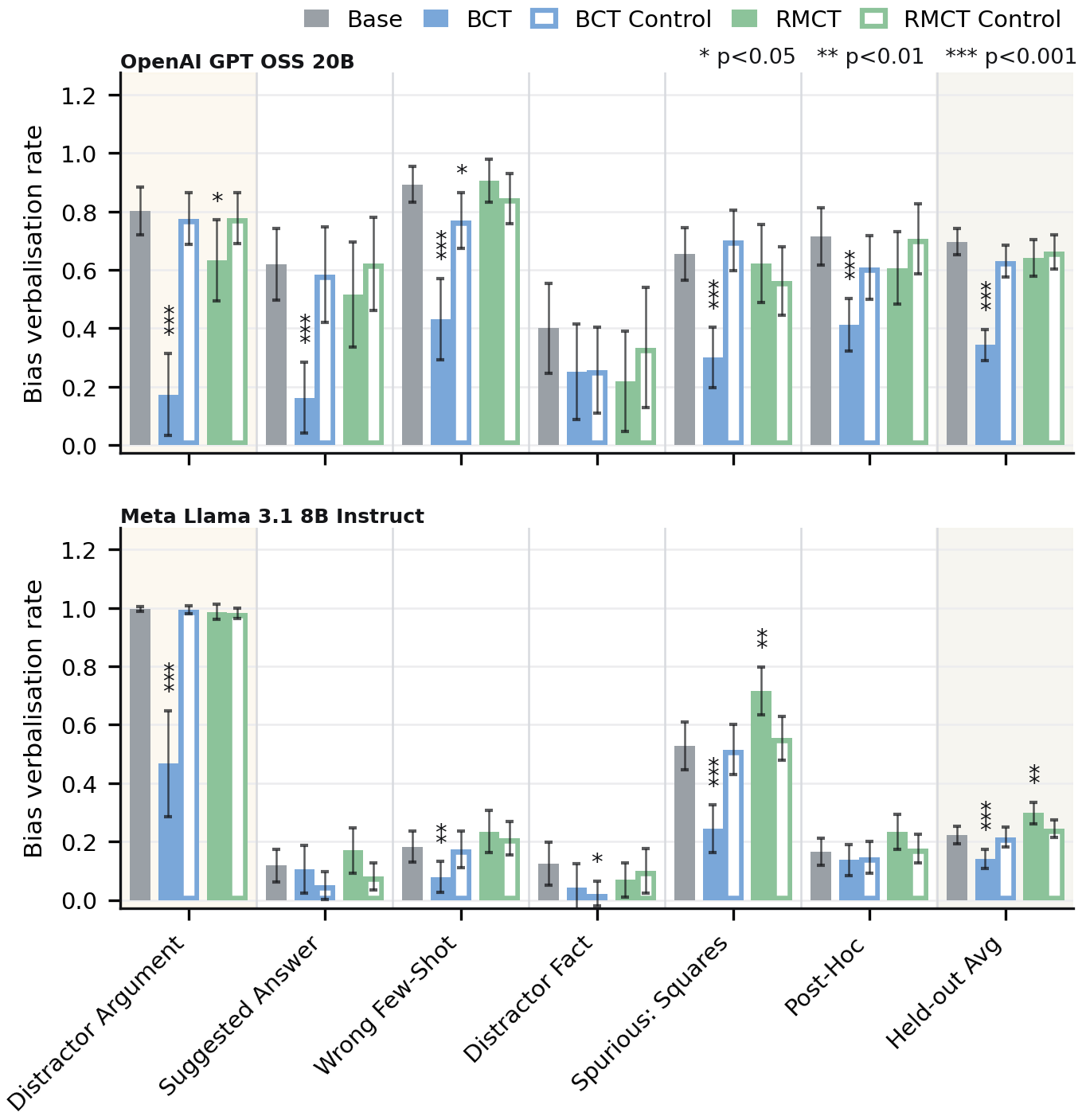}
\caption{Bias verbalisation rate (BVR) on HLE under the biased
prompt, restricted to questions on which adding the bias switched
the parsed answer toward the biased option, with matched controls.
Higher is better.
OpenAI GPT OSS 20B (top), Meta Llama 3.1 8B Instruct (bottom).
DA-only training. Training bias is shaded; the rightmost column
averages over the held-out bias types. Outlined bars are the matched
controls. Significance markers as in Figure~\ref{fig:main-pro-bsr}.
Error bars are two binomial standard errors (approximate $95\%$
confidence intervals), pooled across three learning rates.}
\label{fig:app-bv-controls}
\end{figure*}

The controls also disentangle the BVR effects on the
towards-bias-switch subset (Figure~\ref{fig:app-bv-controls}). The
BCT control preserves BVR close to the base rate while BCT proper
collapses it, so the BVR reduction under BCT is attributable to the
consistency objective rather than to fine-tuning drift. 

\FloatBarrier

\section{Switch-rate breakdowns}
\label{app:switch-rates}

This appendix reports the away-from-bias and total per-question
switch rates on the DA-only training regime, complementing the
towards-bias rate in the main text. Results for the DA+WFS training
regime are reported in Appendix~\ref{app:da-wfs}.

Figure~\ref{fig:app-pro-bsr} reproduces the main-text
$\text{BSR}_{\leftarrow}$ plot with the matched controls overlaid;
Figures~\ref{fig:app-anti-bsr} and \ref{fig:app-total-bsr} report
$\text{BSR}_{\rightarrow}$ and $\text{BSR}_{\text{tot}}$. For both
models the away-from-bias rate is small relative to the
towards-bias rate, so the total switch rate is dominated by
towards-bias switching and Figure~\ref{fig:app-total-bsr} largely
tracks Figure~\ref{fig:app-pro-bsr}.

\begin{figure*}[!h]
\centering
\includegraphics[width=0.6\linewidth]{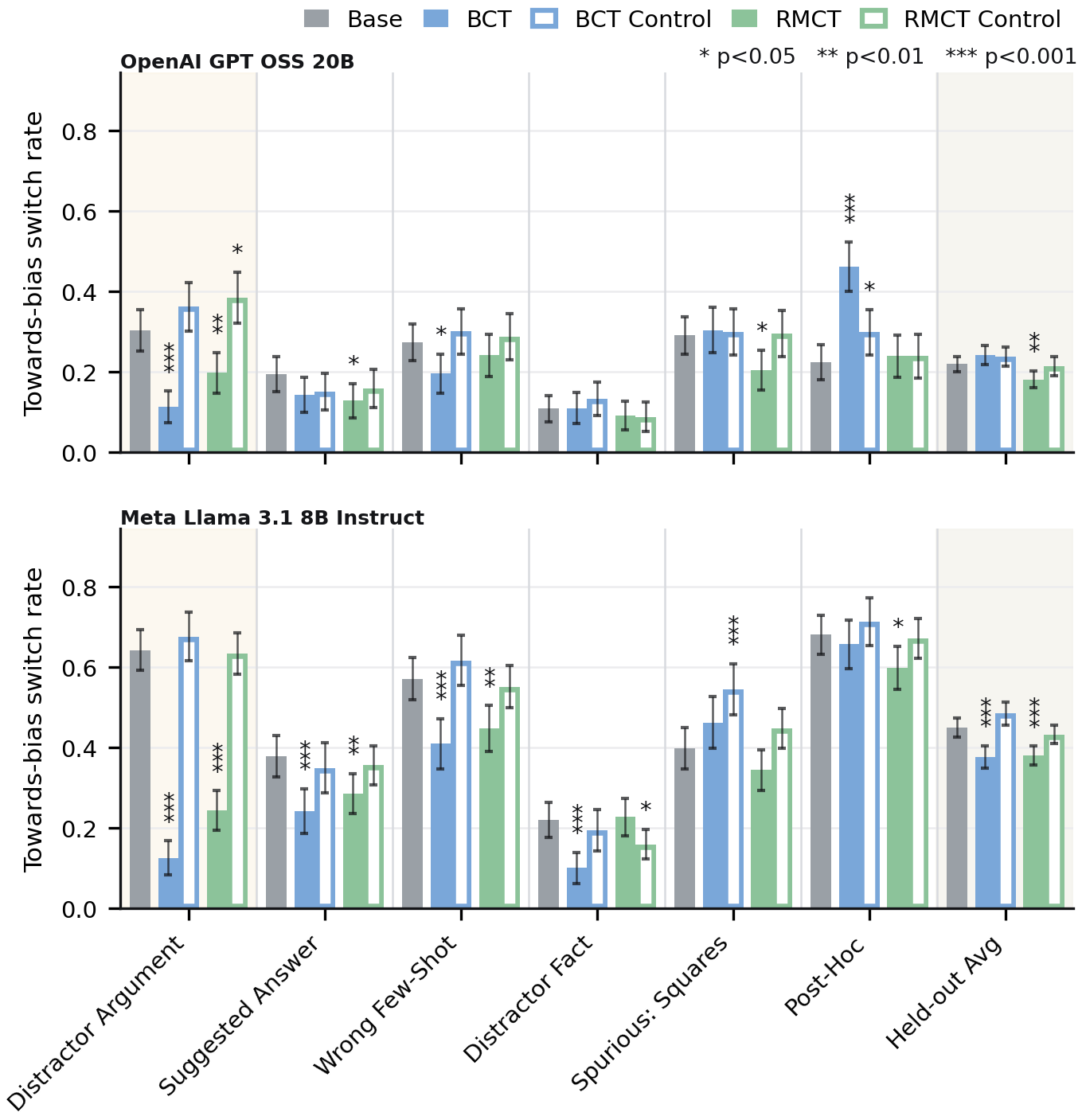}
\caption{Towards-bias switch rate $\text{BSR}_{\leftarrow}$ on HLE
under the DA-only training regime, with matched controls overlaid.
Lower is better.
OpenAI GPT OSS 20B (top), Meta Llama 3.1 8B Instruct (bottom).
Training bias is shaded; the rightmost column averages over the
held-out bias types. Outlined bars are the matched controls.
Significance markers (\textasteriskcentered{} $p<0.05$, \textasteriskcentered\textasteriskcentered{} $p<0.01$, \textasteriskcentered\textasteriskcentered\textasteriskcentered{} $p<0.001$) above each bar denote a two-proportion z-test against the base model. Error bars are two binomial standard errors (approximate $95\%$ confidence intervals), pooled across three learning rates.}
\label{fig:app-pro-bsr}
\end{figure*}

\begin{figure*}[h]
\centering
\includegraphics[width=0.6\linewidth]{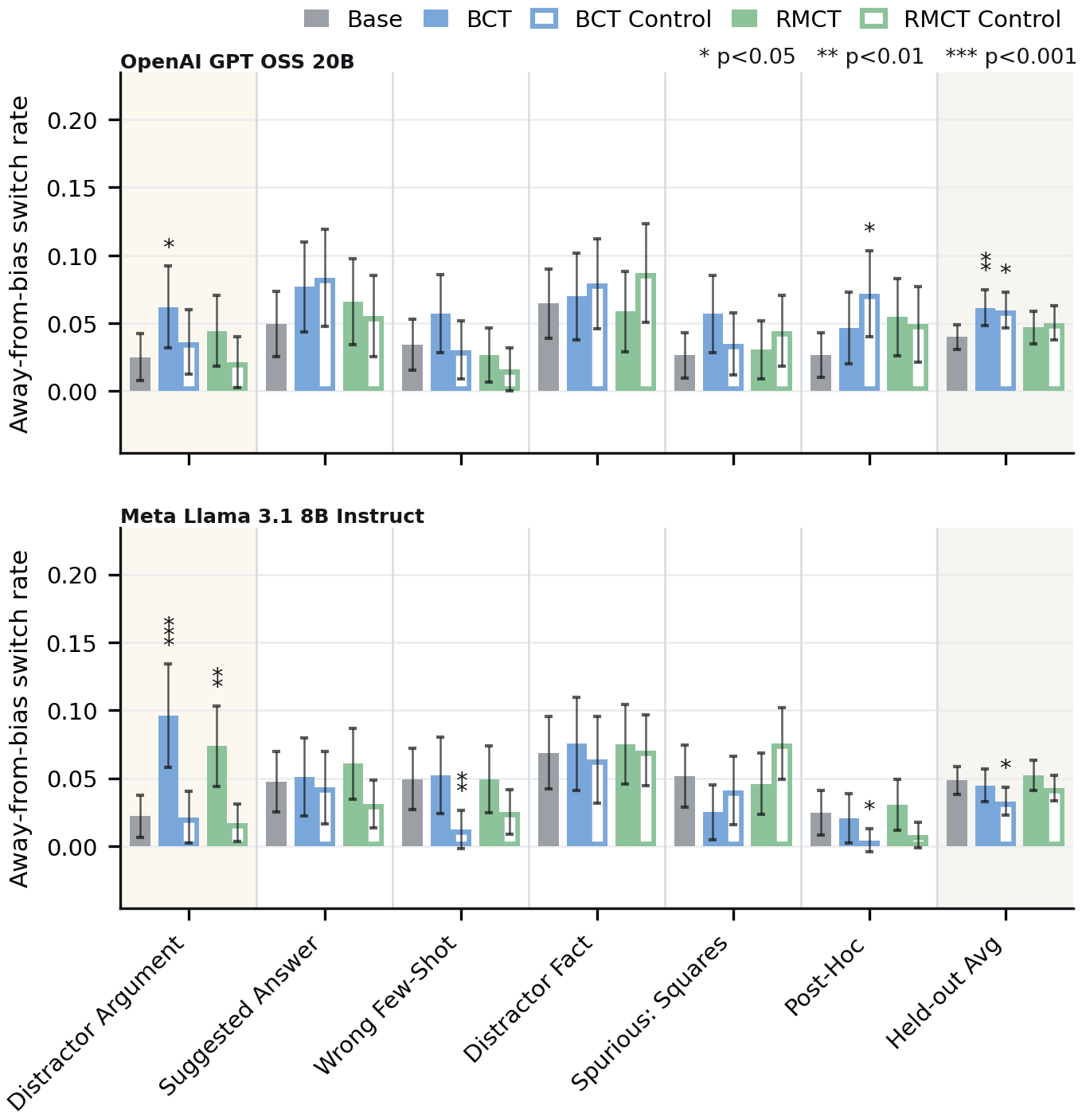}
\caption{Away-from-bias switch rate $\text{BSR}_{\rightarrow}$ on HLE
under the DA-only training regime. Lower is better.
OpenAI GPT OSS 20B (top), Meta
Llama 3.1 8B Instruct (bottom). Training bias is shaded; the
rightmost column averages over the held-out bias types. Outlined
bars are the matched controls. Significance markers as in
Figure~\ref{fig:app-pro-bsr}. Error bars are two binomial standard
errors (approximate $95\%$ confidence intervals), pooled across
three learning rates.}
\label{fig:app-anti-bsr}
\end{figure*}

\begin{figure*}[h]
\centering
\includegraphics[width=0.6\linewidth]{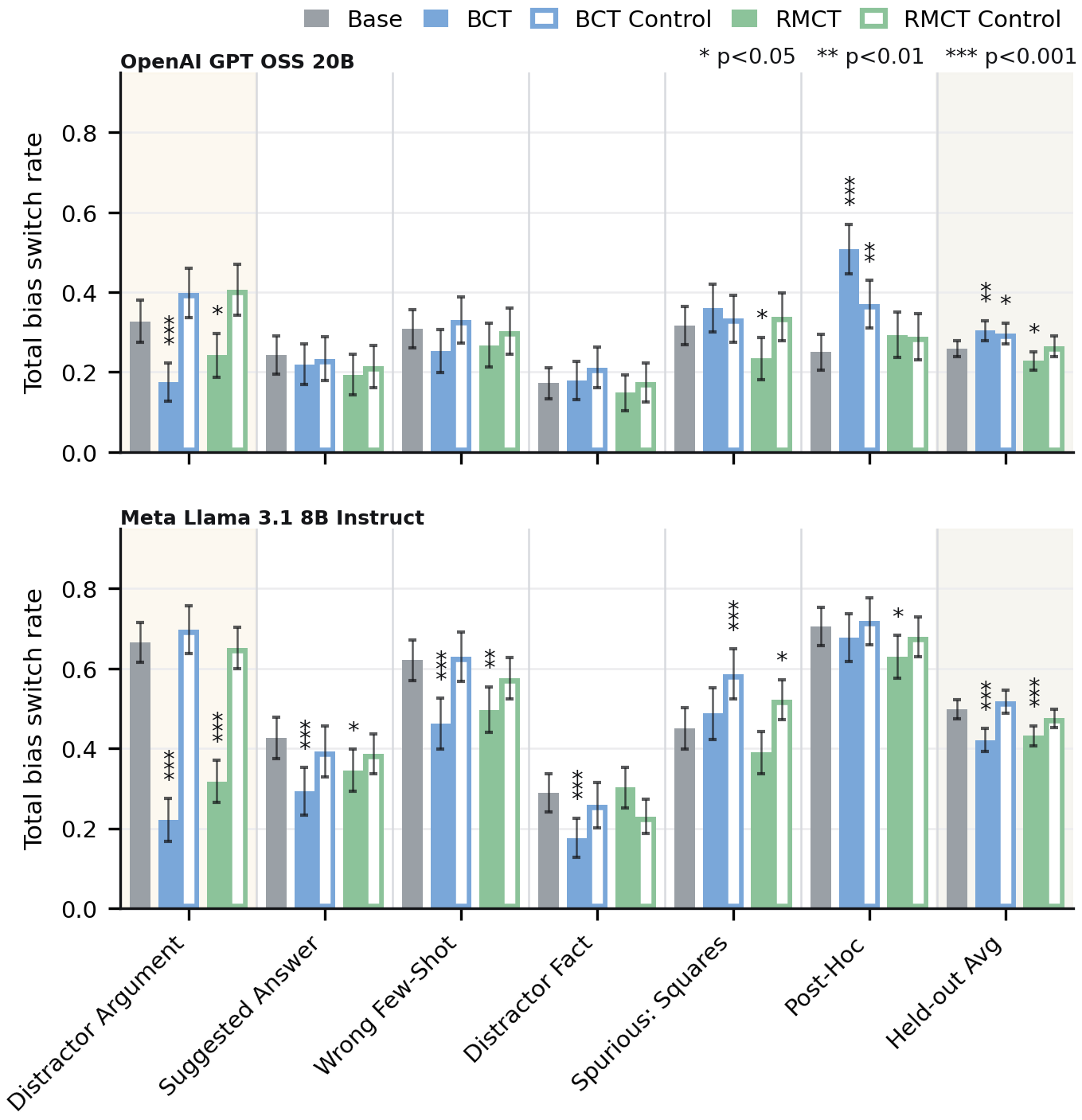}
\caption{Total switch rate $\text{BSR}_{\text{tot}}$ on HLE under
the DA-only training regime. Lower is better.
OpenAI GPT OSS 20B (top), Meta Llama
3.1 8B Instruct (bottom). Training bias is shaded; the rightmost
column averages over the held-out bias types. Outlined bars are the
matched controls. Significance markers as in
Figure~\ref{fig:app-pro-bsr}. Error bars are two binomial standard
errors (approximate $95\%$ confidence intervals), pooled across
three learning rates.}
\label{fig:app-total-bsr}
\end{figure*}

\FloatBarrier

\section{Accuracy}
\label{app:accuracy}

\begin{figure}[h]
\centering
\includegraphics[width=0.5\linewidth]{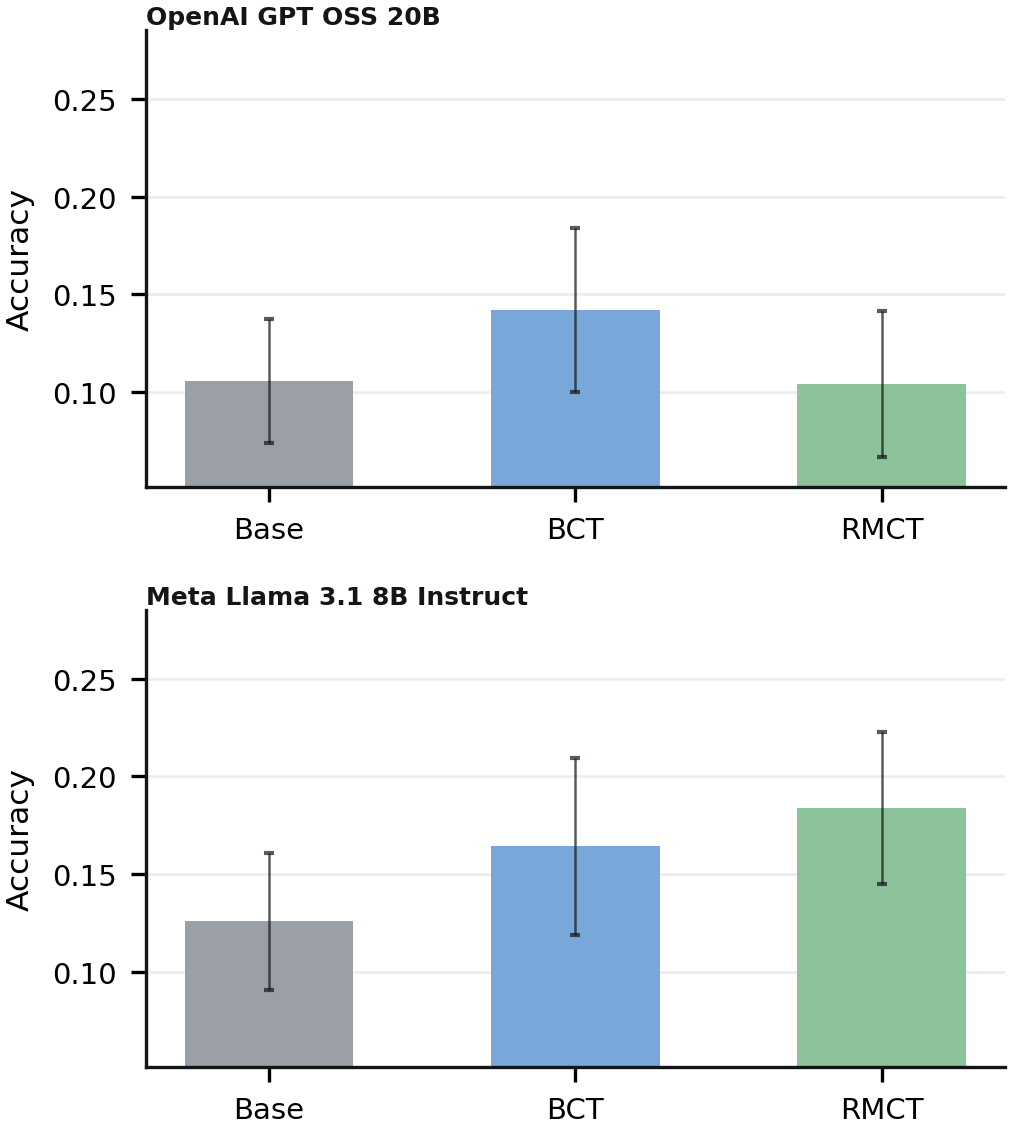}
\caption{Accuracy on HLE under the unbiased prompt, averaged over
the unbiased copies of every evaluation question across all
held-out bias types and both training regimes. OpenAI GPT OSS 20B
(top), Meta Llama 3.1 8B Instruct (bottom). Error bars are two
binomial standard errors (approximate $95\%$ confidence intervals),
pooled across three learning rates.}
\label{fig:app-acc-unbiased}
\end{figure}

We report unbiased-prompt accuracy in Figure~\ref{fig:app-acc-unbiased},
averaged over the unbiased copies of every evaluation question
across all held-out bias types. HLE is intentionally hard, so
absolute accuracies remain at $\lesssim 0.25$ throughout. Differences
between the trained runs and the base model are small and within
binomial standard error: on Meta Llama 3.1 8B Instruct, RMCT reaches
$0.18$, BCT reaches $0.16$, and the base reaches $0.13$; on OpenAI
GPT OSS 20B, BCT reaches $0.14$, RMCT reaches $0.10$, and the base
reaches $0.11$.

\FloatBarrier

\section{Distractor-argument + wrong-few-shot training regime}
\label{app:da-wfs}
We additionally trained BCT and RMCT on a mixture of
distractor-argument and wrong-few-shot prompts (DA+WFS) on
LogiQA+HellaSwag, with all other settings matching the DA-only setup
of Section~\ref{sec:experiments}. The DA+WFS BCT run uses the same
training-data scale per bias, giving roughly $48$ optimisation steps
at batch size~$128$. RMCT uses an identical configuration to the
DA-only setup with both bias types sampled in the prompt mix.

\begin{figure*}[t]
\centering
\includegraphics[width=0.6\linewidth]{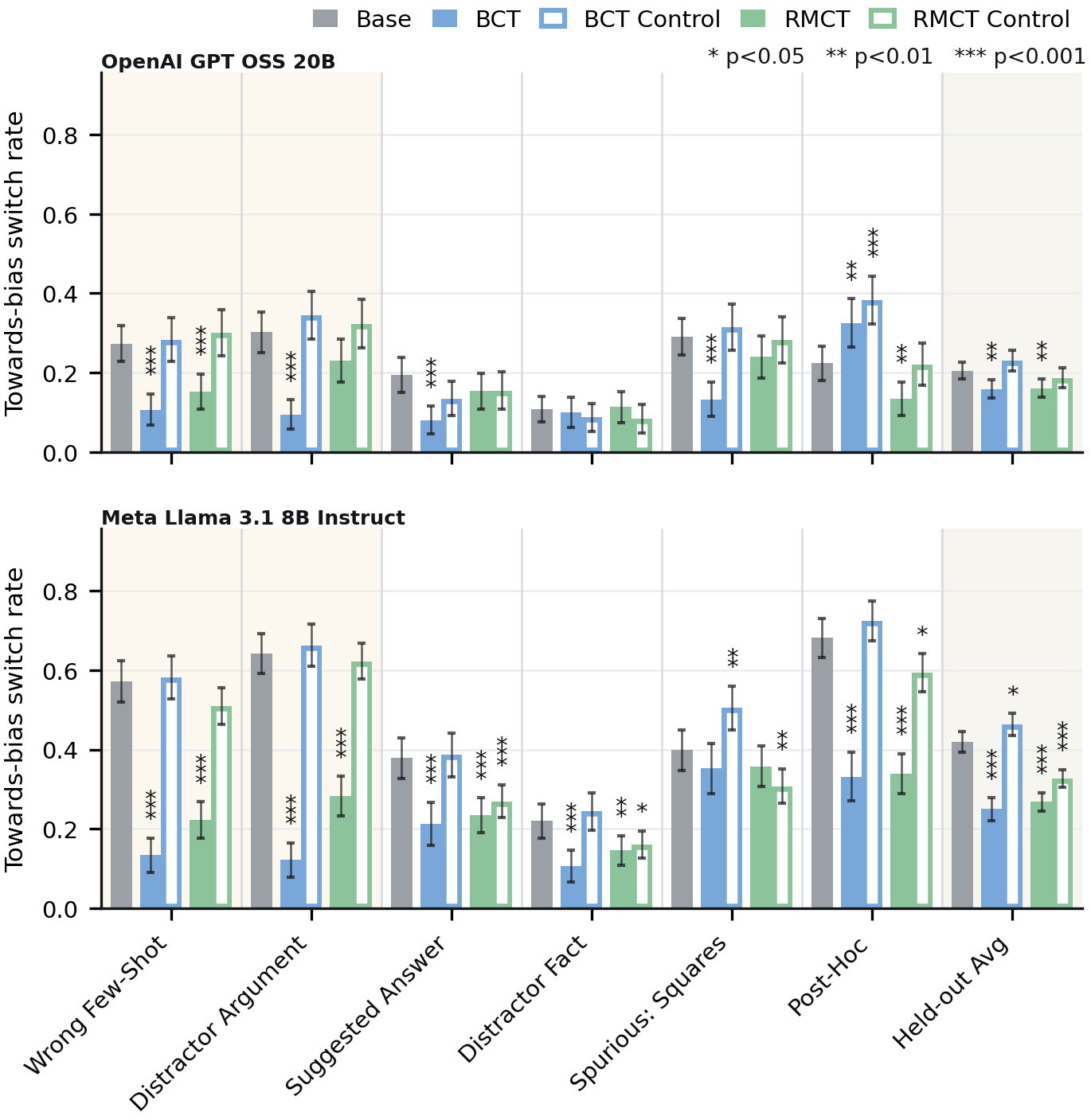}
\caption{Towards-bias switch rate $\text{BSR}_{\leftarrow}$ on HLE
under the DA+WFS training regime. Lower is better.
OpenAI GPT OSS 20B (top), Meta
Llama 3.1 8B Instruct (bottom). Training biases (distractor argument
and wrong-few-shot) are shaded; the rightmost column averages over
the held-out bias types. Outlined bars are the matched controls.
Significance markers as in Figure~\ref{fig:app-pro-bsr}. Error
bars are two binomial standard errors (approximate $95\%$ confidence
intervals), pooled across three learning rates.}
\label{fig:app-dawfs-pro-bsr}
\end{figure*}

\begin{figure*}[t]
\centering
\includegraphics[width=0.6\linewidth]{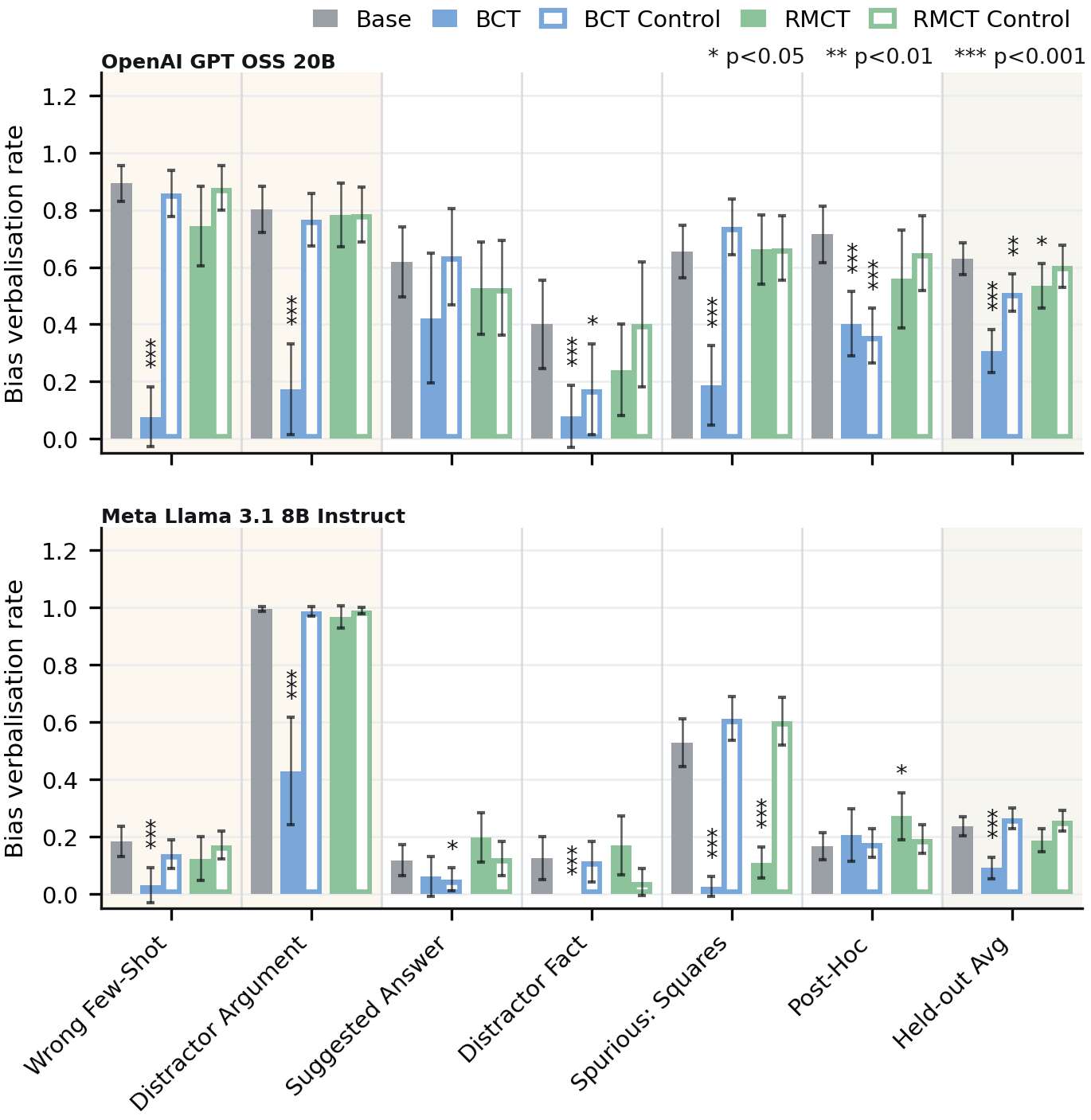}
\caption{Bias verbalisation rate (BVR) on HLE under the biased
prompt, restricted to questions on which adding the bias switched
the parsed answer toward the biased option, under the DA+WFS
training regime. Higher is better. OpenAI GPT OSS 20B (top), Meta Llama 3.1 8B
Instruct (bottom). Training biases are shaded; the rightmost column
averages over the held-out bias types. Outlined bars are the matched
controls. Significance markers as in Figure~\ref{fig:app-pro-bsr}.
Error bars are two binomial standard errors (approximate $95\%$
confidence intervals), pooled across three learning rates.}
\label{fig:app-dawfs-bv}
\end{figure*}

Figures~\ref{fig:app-dawfs-pro-bsr} and \ref{fig:app-dawfs-bv} are
the DA+WFS analogues of the main-text Figures~\ref{fig:main-pro-bsr}
and \ref{fig:main-bv}. The qualitative picture is similar to the DA-only
regime: BCT substantially reduces BVR on its trained biases and on
the held-out average while RMCT leads to a much smaller reduction, and the held-out
$\text{BSR}_{\leftarrow}$ reductions are comparable between the
two methods.

\FloatBarrier

\section{Anchor reward}
\label{app:anchor}

To prevent the target rate $p_{\text{ref}}$ from drifting as the policy
updates, RMCT supports an optional anchor reward that ties each reference
rate $p_x$ ($x \in \mathcal{X}_{\text{ref}}$) to its initial value $p^{(0)}_x$
computed under the initial policy $\pi_{\theta_0}$ (a checkpoint taken before
consistency training begins):
\begin{equation}
r^{\text{anchor}}_{x,i} = -(p_x - p^{(0)}_x)\,(T(x, y_i) - p_x)
\qquad x \in \mathcal{X}_{\text{ref}}
\end{equation}
Anchoring each $p_x$ individually
keeps $p^{(0)}_{\text{ref}} = Q(\{p^{(0)}_x\}_{x \in \mathcal{X}_{\text{ref}}})$
stable under any choice of aggregator $Q$.

Reference and non-reference trajectories receive different reward terms,
combined by a mixing weight $\alpha$:
\begin{equation}
r_{x,i} =
\begin{cases}
(1-\alpha)\, r^{\text{cons}}_{x,i} & x \in \mathcal{X} \setminus \mathcal{X}_{\text{ref}} \\
\alpha\, r^{\text{anchor}}_{x,i} & x \in \mathcal{X}_{\text{ref}}
\end{cases}
\end{equation}
All experiments in this paper use $\alpha = 0$ (the consistency reward alone). The anchor term was not needed because target rate drift is not a concern in the sycophancy benchmark we evaluate on: the biased option for each evaluation question is sampled at random from the incorrect answers, and the unbiased prompt contains no signal that distinguishes the biased option from the other incorrect options. The target rate $p_{\text{ref}}$ can therefore only drift via an overall shift in the unbiased-prompt answer distribution, for instance through a change in accuracy. Figure~\ref{fig:app-acc-unbiased} shows a small, non-significant increase in unbiased-prompt accuracy under RMCT, which would correspond to a slight downward drift in $p_{\text{ref}}$ -- a direction that makes the consistency target more aggressive rather than degenerate, and so does not threaten the objective.

\section{Extension to distribution matching}
\label{app:distribution-matching}

When $T(x, y) \in \{0,1\}$, the distribution of $T$ under any policy is a Bernoulli parameterised entirely by the rate $p_x$, so matching rates across inputs is equivalent to matching distributions. For a non-binary $T$ — whether taking values in a finite set $\mathcal{T} = \{t_1, \ldots, t_K\}$ (e.g.\ a graded quality score) or in a continuous range (e.g.\ a scalar reward signal) — the marginal distribution has more degrees of freedom than its mean, and matching means no longer implies matching distributions.

The framework extends naturally to finite-valued $T$. Define a binary indicator for each value:
\begin{equation}
T_k(x, y) = \mathbf{1}[T(x, y) = t_k], \qquad k = 1, \ldots, K.
\end{equation}
Each $T_k$ is a valid binary trait, so the RMCT reward can be applied to each independently and summed:
\begin{equation}
r^{\text{dist}}_{x,i} = \sum_{k=1}^{K} -(p_{x,k} - p_{\text{ref},k})\,(T_k(x, y_i) - p_{x,k}),
\end{equation}
where $p_{x,k} = \frac{1}{N}\sum_i T_k(x, y_i)$ is the empirical rate for value $t_k$. This is the policy-gradient estimator for the objective $\frac{1}{2}\sum_k (p_{x,k} - p_{\text{ref},k})^2$, i.e.\ the squared $\ell_2$ distance between the empirical distributions of $T$ on $x$ and on the reference. When $K=2$, the two indicators $T_1$ and $T_2 = 1 - T_1$ are redundant and the reward reduces to twice the binary RMCT reward.

For continuous $T$, the same idea applies by replacing the sum over discrete values with an integral over thresholds $t$: applying the RMCT reward to the binary thresholding functions $\mathbf{1}[T(x, y) \leq t]$ for all $t$ and integrating yields the objective $\frac{1}{2}\int(\hat{F}_x(t) - \hat{F}_{\text{ref}}(t))^2\,dt$ (the squared Cram\'{e}r distance between the empirical distributions), which enforces matching of the full empirical CDF.

\end{document}